# Extreme Sparse Multinomial Logistic Regression: A Fast and Robust Framework for Hyperspectral Image Classification

Faxian Cao[1], Zhijing Yang[1*], Jinchang Ren[2], Wing-Kuen Ling[1]

1  School of Information Engineering, Guangdong University of Technology, Guangzhou, 510006, China; 284814227@qq.com; yzhj@gdut.edu.cn; yongquanling@gdut.edu.cn

2  Department of Electronic and Electrical Engineering, University of Strathclyde, Glasgow, G1 1XW, UK; jinchang.ren@strath.ac.uk

*   Correspondence: yzhj@gdut.edu.cn; Tel.: +86-20-39322438

*Abstract:* **Although the sparse multinomial logistic regression (SMLR) has provided a useful tool for sparse classification, it suffers from inefficacy in dealing with high dimensional features and manually set initial regressor values. This has significantly constrained its applications for hyperspectral image (HSI) classification. In order to tackle these two drawbacks, an extreme sparse multinomial logistic regression (ESMLR) is proposed for effective classification of HSI. First, the HSI dataset is projected to a new feature space with randomly generated weight and bias. Second, an optimization model is established by the Lagrange multiplier method and the dual principle to automatically determine a good initial regressor for SMLR via minimizing the training error and the regressor value. Furthermore, the extended multi-attribute profiles (EMAPs) are utilized for extracting both the spectral and spatial features. A combinational linear multiple features learning (MFL) method is proposed to further enhance the features extracted by ESMLR and EMAPs. Finally, the logistic regression via the variable splitting and the augmented Lagrangian (LORSAL) is adopted in the proposed framework for reducing the computational time. Experiments are conducted on two well-known HSI datasets, namely the Indian Pines dataset and the Pavia University dataset, which have shown the fast and robust performance of the proposed ESMLR framework.**

## 1. Introduction

Although the rich spectral information available in a hyperspectral image (HSI) allows classifying among spectrally similar materials [1], supervised classification of HSI remains a challenging task, mainly due to the fact that unfavorable ratio between the limited number of training samples and the large number of spectral band [2]. It may result in the Hughes phenomenon when the spectral bands increase and leads to poor classification results [3]. To tackle such challenges, a number of state-of-the-art techniques have been proposed, such as the support vector machine (SVM) [4], the multi-kernel classification [5], the extreme learning machine (ELM) [6] and the sparse multinomial logistic regression [2, 7-10] (SMLR). In addition, many approaches have also been proposed for dimensionality reduction and feature extraction [11-12], which include the principal component analysis and its variations [13-16] (PCA), the extended multi-attribute profiles [17-18] (EMAPs), the singular spectrum analysis [19-22] (SSA) and the segmented auto-encoder [12, 22]. Among these methods, the SMLR [23-24] has drawn a lot of attentions due to its good performance.

The SMLR has been proved to be robust and efficient under the Hughes phenomenon and is able to learn the class distributions in a Bayesian framework [25]. Hence, it can provide a degree of plausibility for performing these classifications [26]. Moreover, the logistic regression via the variable splitting and the augmented Lagrangian (LORSAL) has been proposed for dealing with large datasets and multiple classes efficiently. Since it can effectively learn a sparse regressor with a Laplacian prior distribution of the SMLR, the combination of SMLR and LORSAL is found to be one of the most effective methods for coping with the high dimensional data of HSI [26-27].

However, the existing SMLR framework suffers from some severe drawbacks. First, the SMLR with the original spectral data of the HSI as features is inefficient, thus it is necessary to find a better representation of the HSI data for improved classification accuracy. The second is the manually set initial value for the regressor, which may result in poor classification of HSI due to improper initial value used. Recently, some deep learning algorithms such as the convolutional neural network [28-29] (CNN) and the extreme learning machine (ELM) have drawn lots of attentions due to their good classification results for the HSI [30-32]. However, CNN requires huge computational time and seem unrealistic. ELM is a generalized single layer feedforward neural networks (SLFNs), which characterizes fast implementation, strong generalization capability and a straightforward solution [6]. The main goals of ELM are to minimize the output weights of the hidden layer and maintain the fast speed. According to the Bartlett's neural network generalization theory [33], the smaller norm of the weights will lead to better generalized performance. Hence, a feedforward neural network can reach a smaller training error [34].

For efficiency, the input weights and the bias between the input layer and the hidden layer of the ELM are randomly generated. It has been proved to be a fast and good data representation method [30-32]. In fact, besides ELM, some other models, such as the liquid state machines [35] and the echo state networks [36-37] have also adopted the random weight selection technique with great success [38]. Therefore, the problem of the poor data representation in the SMLR can be addressed using the random weight selection technique. Hence, in this paper, we propose the extreme sparse multinomial regression (ESMLR) for the classification of HSI. First, the data in the HSI will be represented by randomly generated weight and bias for SMLR, which also maintain the fast speed of the SMLR and improve the representation performance. Second, we set up an optimization model to minimize the training error of the regressor value, which is solved by using the Lagrange multiplier method and dual principle in order to automatically find a better initial regressor value for the SMLR (detailed in Section III).

In addition to spectral features, spatial information is also very important for the classification of the HSI. In the proposed ESMLR framework, the extended multi-attribute profile (EMAP) is used for feature extraction, as both morphological profiles (MPs) [39] and the attribute profiles (APs) [16-17] have been successfully employed for performing the spectral and spatial HSI classification. Moreover, the linear multiple feature learning (MFL) [7] is employed to maintain the fast speed and further improve the classification results. The MFL has been proposed for adaptively exploiting the information from both the derived linear and nonlinear features. As a result, it can potentially deal with the practical scenarios that different classes in the HSI datasets need different (either nonlinear or linear) strategies [7]. According to the Li's works [7], the nonlinear feature such as the kernel feature contributes little to the HSI classification when the MFL is utilized. Moreover, it requires much more computational efforts for processing the nonlinear features. Therefore, a linear combination of the MFL which just utilizes the linear features of the HSI is proposed for the ESMLR. Hence, this operation can not only improve the classification results, but also maintain the fast speed of the ESMLR.

The main contributions of the proposed ESMLR framework in this paper can be highlighted as follows. First, the problem of the SMLR that uses the initial data of the HSI for performing the classification is addressed by randomly generating the input weights and bias of the input data, which will not only maintain the fast processing speed, but also improve the classification results of the HSI. Second, a new principle is introduced to automatically determine a suitable initial regressor value for SMLR to replace the manually settings. Third, the linear combination of the MFL that integrates the spectral and spatial information of HSI extracted by EMAPs followed by LORSAL is employed for the ESMLR for fast and robust data classification of HSI.

The remainder of this paper is structured as follows. Section II describes the background of the SMLR and the EMAPs. In Section III, the proposed ESMLR framework is presented. The experiment results and discussions are summarized in Section IV. Finally, Section V concludes this paper with some remarks and suggestions for the plausible futures.

## 2. Proposed Classification Framework

### 2.1. EMAPs

The APs are obtained by applying attribute filters (AFs) to a gray-level image [16]. The AFs are connected operators defined via a mathematical morphological mean for a gray level image to keep or to merge their connected components at different gray levels [39]. Let $\gamma$ and $\phi$ be an attribute thinning and an attribute thickening based on an arbitrary criterion $T_\lambda$. Given an image $f_i$ and a sequence of thresholds $\{\lambda_1, \lambda_2, \ldots, \lambda_p\}$, an AP can be obtained by applying a sequence of attribute thinning and attribute thickening operations as follows:

$$AP(f_i) = \{\phi_{\lambda_p}(f_i), \phi_{\lambda_{p-1}}(f_i), \ldots, \phi_{\lambda_1}(f_i), f_i, \gamma_{\lambda_1}(f_i), \ldots, \gamma_{\lambda_{p-1}}(f_i), \gamma_{\lambda_{p-1}}(f_i)\}. \qquad (1)$$

Note that in (1) the AP is defined on each spectral band, hence the dimensionality of the APs will be very high when it is applied for the full spectral bands of the HSI [39]. In [40], the principal component analysis (PCA) was suggested to solve this problem. Hence, the extended AP (EAP) is acquired by generating an AP on each of the first c PCs below [41].

$$EAP = \{AP(PC_1), AP(PC_2), \ldots, AP(PC_c)\} \qquad (2)$$

Then, the EMAPs is defined as the composition of b different EAPs based on a set of b attributes $\{a_1, a_2, \ldots, a_b\}$ as follows:

$$EMAPs = \{EAP_{a_1}, EAP_{a_2}, \ldots, EAP_{a_b}\}. \qquad (3)$$

Although a wide variety of attributes can be applied to the APs [42] for performing the HSI classification, in this paper we only consider the area attribute in order to maintain the fast speed whilst incorporating the spectral and spatial information. Here, the code of the APs is from online http://www.lx.it.pt/~jun/. The threshold values of the area attribute were chosen as 100, 200, 500 and 1000. The first c PCs are determined to have the cumulative eigenvalues larger than 99% of the total value.

*2.2. SMLR*

Let t={1, …, M} be a set of M class labels. Denote S={1, …, n} as a set of integers indexing the n pixels of any HSI and x=$(x_1, \ldots, x_N) \in R^{d \times N}$ be the HSI. Here, each pixel in the HSI is a d-dimensional feature vector and $y = (y_1, \ldots, y_N)$ denote the labels of x. Let $D_n = \{(x_1, y_1), \ldots, (x_L, y_n)\}$ be the training set. All the above parameters will be discussed in Section III.

First of all, the posterior class probabilities are modeled by the MLR [2], [7] as follows:

$$p(y_i = m | x_i, w) := \frac{exp\left(w^{(m)^T} h(x_i)\right)}{\sum_{m=1}^{M} exp\left(w^{(m)^T} h(x_i)\right)}, \qquad (4)$$

where $w = [w^{(1)}, \ldots, w^{(M-1)}]^T \in R^{M-1 \times d}$ denotes the regressors and $h(x_i)$ denotes the input feature. Here, the superscript 'T' denotes the transpose operator of a matrix. $w^M$ is set to be 0 because the densities of (4) do not depend on the translation of the regressor $w^M$ [7]. The input features h can be linear or nonlinear. In the former case, we have:

$$h(x_i) = [x_{i1}, \ldots, x_{id}]^T, \qquad (5)$$

where $x_{i,j}$ is the j-th component of $x_i$.

If $h(.)$ is nonlinear, it can be formulated as follows:

$$h(x_i) = [1, \varphi_1(x_i), \ldots, \varphi_d(x_i)]^T, \qquad (6)$$

where $\varphi(.)$ is a nonlinear function.

According to [2], [7], the regressor w of the SMLR can be obtained by calculating the maximum a posteriori estimate as follows:

$$\hat{w} = \arg \max_w \{\ell(w) + \log(p(w))\}, \qquad (7)$$

Here, $\ell(w)$ is the logarithmic likelihood function given by:

$$\ell(w) := \log \prod_{i=1}^{n} p(y_i | x_i, w) = \sum_{i=1}^{n} \left(h^T(x_i) w^{(y_i)} - \log \sum_{m=1}^{M} exp\left(h^T(x_i) w^{(m)}\right)\right), \qquad (8)$$

It is worth noting that $\log p(w)$ is a prior over w which is irrelevant to the observation x. For controlling the complexity and the generalization capacity of the classifier, w is modeled as a random vector with the Laplacian density denoted as

$$p(w) \propto exp(-\lambda \|w\|_1)$$

Here, $\lambda$ is the regularization parameter controlling the degree of sparsity [2]. Hence, the solution of SMLR can be expressed as follows:

$$\hat{w} = \arg \max_w \{\sum_{i=1}^{n} \left(h^T(x_i) w^{(y_i)} - \log \sum_{m=1}^{M} exp\left(h^T(x_i) w^{(k)}\right)\right) + \log p(w)\} \qquad (9)$$

The LORSAL algorithm is applied to SMLR to cope with the larger size problem of the HSI data.

*2.3. From SMLR to ESMLR*

The MLR can be modeled as follows [24], [43]:

$$p(y_i = m|x_i, w) = \frac{exp(w^{(m)T}h(x_i))}{1+\sum_{m=1}^{M-1} exp(w^{(m)T}h(x_i))} \quad (10)$$

and

$$p(y_i = M|x_i, w) = \frac{1}{1+\sum_{m=1}^{M-1} exp(w^{(m)T}h(x_i))} = 1 - \sum_{m=1}^{M-1} p(y_i = m|x_i, w), \quad (11)$$

where $h(x_i)$ is the input feature of the MLR and $w = [w^{(1)}, \dots, w^{(M-1)}]^T \in R^{(M-1)\times d}$ denotes the regressors. $w^M$ is set to be 0, as the densities of (10) and (11) do not depend on the translation of the regressor $w^M$ [7]. The input features h can be linear or nonlinear.

If the input feature is linear, then we have:

$$h(x_i) = [x_{i,1}, \dots, x_{id}]^T, \quad (12)$$

where $x_{i,j}$ is the j-th component of $x_i$.

If h is nonlinear, it can be formulated as:

$$h(x_i) = [1, \varphi_1(x_i), \dots, \varphi_d(x_i)]^T, \quad (13)$$

where $\varphi(.)$ is a nonlinear function.

The initial regressor value can be used to find a better representation of the HSI for the ESMLR via determining the solution of the following optimization problem:

$$\text{Minimize } ||wH - Y||^2 \text{ and } ||w||^2, \quad (14)$$

where $Y = [y_1^*, \dots, y_n^*] \in R^{(M-1)\times n}$, and

$$H = \begin{pmatrix} h(a_1, b_1, x_1) & \cdots & h(a_1, b_1, x_n) \\ \vdots & \ddots & \vdots \\ h(a_L, b_L, x_1) & \cdots & h(a_L, b_L, x_n) \end{pmatrix},$$

$$h_{spa}(a_i, b_i, x_i) = \frac{1}{1+exp(-(a_i^T x_i + b_i))}.$$

Here, if $x_i$ belongs to the j-th class, $y_{i,j}^* = 1$. Otherwise, $y_{i,j}^* = 0$. In fact, the activation function $h()$ can be either linear or nonlinear, and L is the dimension of the feature space which we want to project; $a_i \in R^d$ and $b_i \in R^1$ are randomly generated. Actually, a wide range of feature mapping functions can be considered in our work which include but not limit to:

(1) Linear function: $h(a_i, b_i, x_i) = a_i^T x_i + b_i$;

(2) Sigmoid function: $h(a_i, b_i, x_i) = \frac{1}{1+exp(-(a_i^T x_i + b_i))}$;

(3) Gaussian function: $h(a_i, b_i, x_i) = exp(-b_i \| a_i^T x_i \|^2)$;

(4) Hardlimit function: $h(a_i, x_i) = \begin{cases} 1 & \text{if } a_i^T x_i \geq 0 \\ 0 & \text{otherwise} \end{cases}$;

(5) Multiquadrics function $h(a_i, b_i, x_i) = (\| x_i - a_i \|^2 + b_i^2)^2$, etc.

From (14), it can be seen that the objective of the optimization is not only to reach a smaller training error, but also to reach a smaller value of the regressor w. According to the Bartlett's theory [33], this will help the proposed approach to achieve a good performance. From the optimization theory viewpoint [30]-[34], (14) can be reformulated as follows:

$$\min_w \frac{1}{2}||w||_F^2 + C \frac{1}{2}\sum_{i=1}^n ||\xi_i||_2^2, \text{ s.t. } wh(x_i) = y_i^T - \xi_i^T \text{ for i=1,...,n}, \quad (15)$$

where C is a regularization parameter and $\xi_i$ is the training error for the samples $x_i$.

Based on the Karush Kuhn Tucker optimality conditions and the Lagrange multiplier method [44], we have:

$$L_{ESMLR} = \frac{1}{2} * \| w \|_F^2 + C * \frac{1}{2}\sum_{i=1}^n ||\xi_i||_2^2 - \sum_{i=1}^n \sum_{j=1}^{M-1} \alpha_{ij} (w_j h(x_i) - y_{ij}^T + \xi_{ij}^T), \quad (16)$$

where $\alpha_{i,j}$ is the Lagrange multiplier.

Then, the optimization condition can be expressed as follows:

$$\frac{\partial L_{ESMLR}}{\partial w} = 0 \rightarrow w = \alpha H^T, \quad (17)$$

$$\frac{\partial L_{ESMLR}}{\partial \xi_i} = 0 \to \alpha_i = C\xi_i, \tag{18}$$

$$\frac{\partial L_{ESMLR}}{\partial \alpha_i} = 0 \to wh(x_i) - y_i^T + \xi_i^T \text{ for } i = 1, \dots, n, \tag{19}$$

where $\alpha_i = [\alpha_{i,1}, \alpha_{i,2}, \dots, \alpha_{i,M-1}]^T$ and $\alpha = [\alpha_1, \alpha_2, \dots, \alpha_n]$.

Hence, the solution of the optimization defined in (15) can be analytically expressed as

$$w = H(\frac{I}{C} + H^T H)^{-1} Y^T, \tag{20}$$

$$\text{or} \quad w = (\frac{I}{C} + HH^T)^{-1} HY^T. \tag{21}$$

From (20) and (21), it can be seen that the initial regressor value $w_0$ is good for the ESMLR satisfying the optimization condition. Here, the random weight function h() can be used not only to find a better representation of the HSI data, but also to maintain the fast speed for the proposed framework. Based on the principles of the SMLR algorithm [23], the regressor $w$ of the proposed ESMLR at the k-th iteration can be computed by the maximum a posterior estimate as follows:

$$\widehat{w}_k = \arg \max_w \ell(w_{k-1}) + \log p(w_{k-1}) \text{ for } k = 1, 2, \dots, \tag{22}$$

where $p(w_{k-1}) \propto \exp(-\lambda \|w_{k-1}\|_1)$ and $\lambda$ is the regularization parameter for controlling the degree of sparsity [23].

The solution of (22) at the k-th iteration can be addressed by introducing the linear or nonlinear input features. That is, for the linear case: $h(x_i) = [h(x_{i,1}), \dots, h(x_{id})]^T$; for the nonlinear (kernel) case: $h(x_i) = [1, \phi_1(x_i, x_1), \dots, \phi_d(x_i, x_d)]^T$, where $\phi$ is a nonlinear function. Also, we have:

$$\widehat{w}_k = \arg \max_w \sum_{i=1}^n (w_{k-1}^{y_i} h(x_i) - \log \sum_{m=1}^{M-1}(1 + \exp(w_{k-1}^m h(x_i)))) + \log p(w_{k-1}). \tag{23}$$

Similar to K-SMLR [2], [26], the proposed ESMLR can also be extended to form a kernel-based ESMLR (K-ESMLR). The performance of the proposed ESMLR and K-ESMLR are evaluated in next section. In order to address the larger size problem of the HSI data, including large datasets and the number of classes, the LORSAL [45] algorithm is adopted. Moreover, the EMAPs are utilized for performing the efficient feature extraction and incorporating the spectral information and the spatial information.

*2.4. ESMLR with a linear MFL*

As mentioned above, the spectral information and the spatial information are integrated to further improve the performance of the proposed framework. It is well known that the kernel transform will increase the size of the input feature. As shown in [2],[7], the kernel transform may contribute slightly on the HSI classification accuracy when nonlinear features are utilized for the MFL. The kernel feature will also slow the speed of algorithms. Based on this perspective, a combinational linear MFL is proposed for improved HSI data classification whilst maintaining the low computational time of the proposed ESMLR.

Let $h_{spe}(x_i)$ and $h_{spa}(x_i)$ be the input spectral features of the raw/original HSI data and the spatial features extracted by the EMAPs, respectively. The input features of the proposed ESMLR can be expressed as follows:

$$h(x_i) = [h_{spe}(x_i), h_{spa}(x_i)]^T. \tag{24}$$

Then, (23) can be reformulated as:

$$\widehat{w}_k = \arg \max_w \log p(w_{k-1}) + \sum_{i=1}^n (w_{k-1}^{y_i} h_{spe}(x_i) + w_{k-1}^{y_i} h_{spa}(x_i)$$

$$- \log \sum_{m=1}^{M-1}(1 + \exp(w_{k-1}^m h_{spe}(x_i) + w_{k-1}^m h_{spa}(x_i))) \tag{25}$$

From (25), it can be seen that (23) and (25) have the same structure. Therefore, the LORSAL algorithm will be adopted in the proposed framework. Figure 1 shows the flowchart of the proposed spectral spatial ESMLR framework.

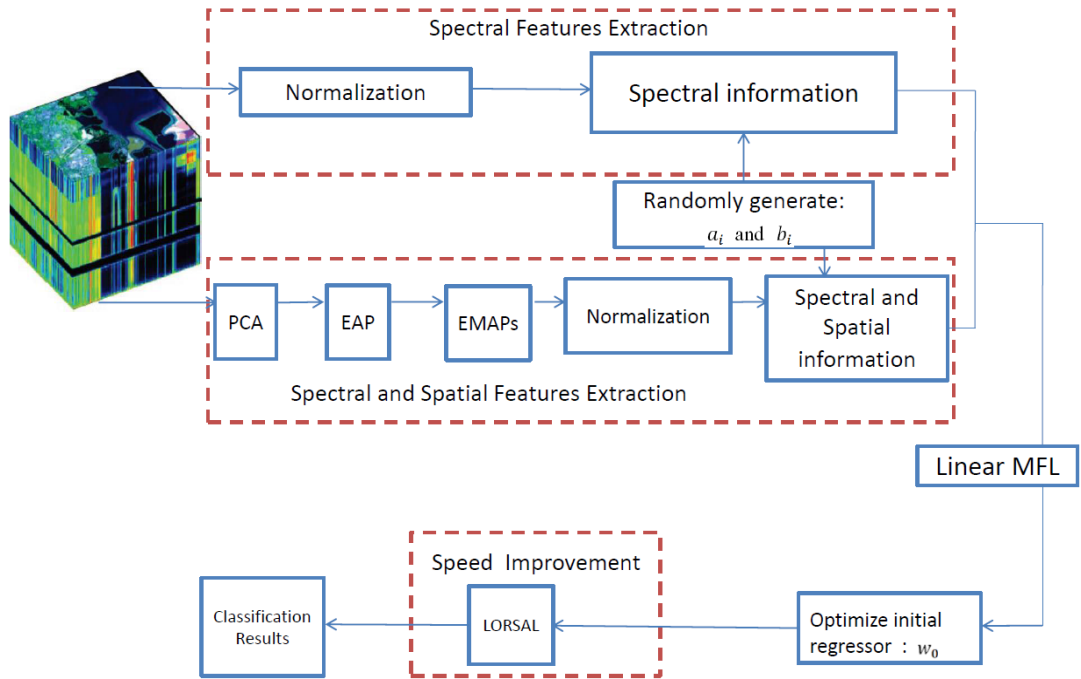

**Figure 1.** The flowchart of the proposed ESMLR framework.

## 3. Experimental Section

In this section, the proposed ESMLR and K-ESMLR will be evaluated and relevant results are summarized in detail as follows.

*3.1. The datasets*

Two well-known HSI datasets are used in our experiments, which are detailed below.

1). The Indian Pines dataset: The HSI image was acquired by the AVRIS sensor in 1992. The image contains 145×145 pixels and 200 spectral bands after removing 20 bands influenced by the atmospheric affection. There are 10366 labelled samples in 16 classes within the HSI dataset.

2). The Pavia University dataset: The system was built by the University of Pavia of Italy in 2001. The Pavia University dataset was acquired by the ROSIS instrument. The image contains 610×340 pixels and 103 bands after discarding 12 noisy and water absorption bands. In total there are 42776 labelled samples in 9 classes within this dataset.

*3.2. Compared Methods and parameter settings*

The proposed ESMR framework are compared with the classical classifiers such as the K-SVM [34] (The codes of the K-SVM are obtained online from http://www.fst.umac.mo/en/staff/fstycz.html/), the SMLR and the K-SMLR [2], [7] (The codes of the SMLR and the K-SMLR are from online http://www.lx.it.pt/~jun/). All experiments are conducted in MATLAB R2015a and tested on a computer i7 with 3.40GHz CPU and 8.0G RAM. All data are normalized via the unit max method, i.e. each data of a HSI is divided by the largest value of the whole dataset.

For all kernel-based/nonlinear methods, the Gaussian radial basis function (RBF) kernel is used. For the parameter σ of the RBF in the K-SMLR and the K-ESMLR, it is set to be 0.85 for the Indian Pines dataset and 0.35 for the Pavia University dataset as suggested by Sun et al [26]. The LIBSVM toolbox of the MATLAB R2015a is used for the implementation of the K-SVM approach [46]. The parameter of the K-SVM were chosen according to [34]. For the cost parameter in (20) or (21), $C = 2^a$ is chosen where a is in the range $\{1, 2, …, 20\}$. The regularization parameter in (22) is set to $\lambda = 2^b$. The total number of dimension of the new feature space L is chosen in the range $\{50, 100, …, 1450, 1500\}$. If there is no special emphasis required, the dimension of the new feature space in the proposed ESMLR is set to be L=300 for spectral information only and the combined spectral-spatial information (EMAPs). Similarly, L=500 is chosen for the situation that utilize spectral and spatial information (linear MFL). For other parameters in the SMLR, K-SMLR, ESMLR and K-ESMLR, details will be discussed in the subsections below. All experiments are repeated 10 times with the average classification results reported for comparison.

*3.3. Discussions on the robustness of the ESMLR framework*

In the following experiments, the robustness of the proposed framework is evaluated. For the Indian Pines dataset and the Pavia University dataset, in total 515 and 3921 samples are randomly selected for training, respectively. The remaining samples are used for testing based on the overall accuracy (OA) of classification. For the two datasets, the number of samples used for training and testing from each class are summarized in Table 1 and Table 2, respectively, along with the classification results under different experiments settings as detailed below.

**Experiment 1**: In this experiment, the proposed ESMLR approach is evaluated in three different situations, i.e. using only the spectral information as features, using both spectral and spatial information yet in combination with EMAPs and the linear MFL, respectively. For the Indian Pines and the Pavia University datasets, the results are shown in Figure 2 and Figure 3, respectively. As seen, the proposed framework shows the good performances in all the situations when L>150. The fusion of spectral and spatial information can successfully improve the OA, where the combination of ESMLR and MFL slightly outperforms ESMLR and EMAPs.

**Experiment 2**: In this experiment, the impact of the parameter C ($C = 2^a$) in the proposed K-ESMLR under the aforementioned three different situations are evaluated. As shown in Figure 4, the proposed ESMLR achieves a very good performance when C is larger than 0, where the classification results are very stable even though a or C is significantly changed. This again demonstrates the robustness of the proposed framework.

**Experiment 3**: The impact of the sparse parameter b ($\lambda = 2^b$) on (22) are evaluated in this experiment. As shown in Figure 5, the proposed K-ESMLR achieves better classification results compared with the SMLR when only spectral information is utilized, especially for the Pavia University dataset. Also it seems K-ESMLR slightly outperforms K-SMLR. This has demonstrated that the proposed framework achieves a better performance compared with the conventional SMLR framework for both linear and nonlinear (kernel) cases. When EMAPs are applied to extract both the spectral and spatial information, the proposed framework also achieves better classification results compared with SMLR. When combining our ESMLR framework with the proposed combinational linear MFL, it also outperforms SMLR. Note that K-SMLR and K-ESMLR cannot be combined with linear MFL, hence the results are not shown. In summary, thanks to the proposed improvements, the proposed ESMLR/K-ESMLR framework has outperformed conventional SMLR and K-SMLR for effective classification of HSI.

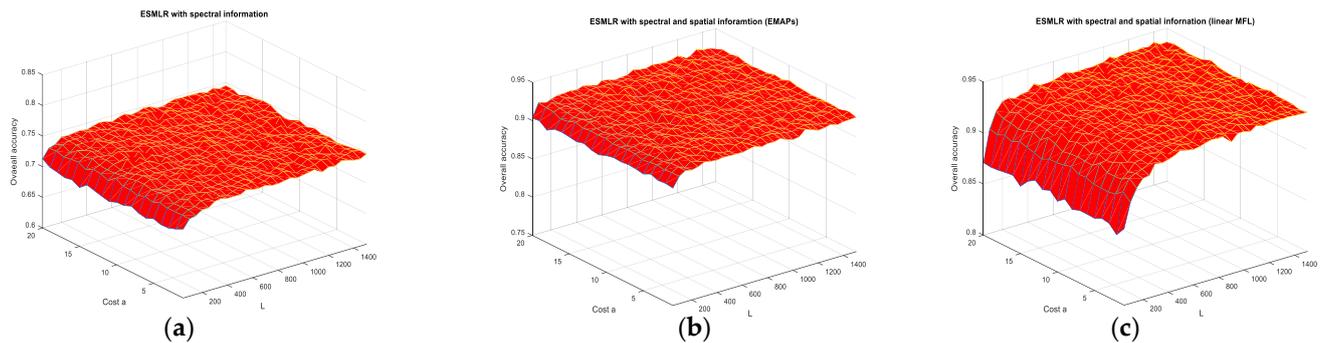

**Figure 2.** The robustness performance of the proposed framework based on the Indian Pines dataset: (**a**) The proposed ESMLR with spectral information (200 features and b=-7); (**b**) The proposed ESMLR with spectral and spatial information (EMAPSs) (36 features and b=-11); (**c**) The proposed ESMLR with spectral and spatial information (proposed linear MFL) (236 features and b=-10).

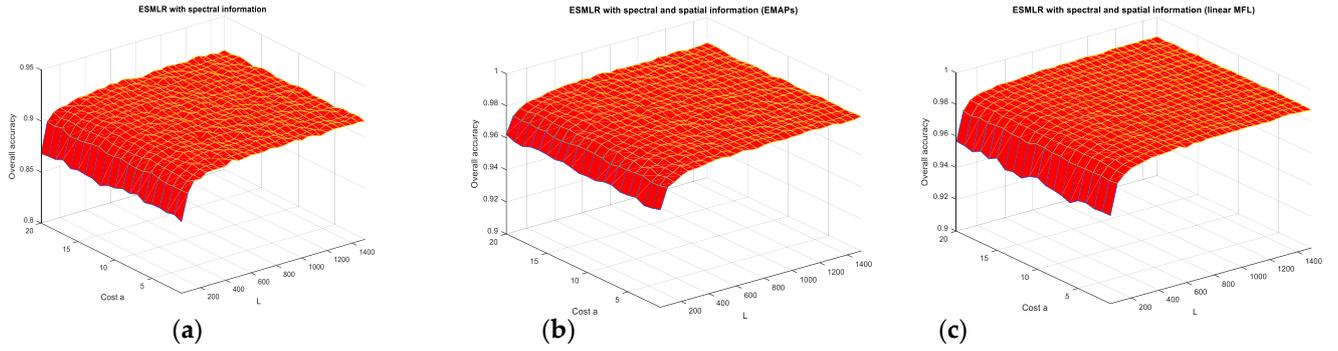

**Figure 3.** The robustness performance of the proposed framework based on the Pavia University dataset: (**a**) The proposed ESMLR with spectral information (200 features and b=-11); (**b**) The proposed ESMLR with spectral and spatial information (EMAPs) (36 features and b=-11); (**c**) The proposed ESMLR with spectral and spatial information (proposed linear MFL) (236 features and b=-11).

*3.4 Discussions on classification results and the running time of different algorithms*

In this subsection, the classification results and the running (executed) time based on the proposed classifiers are compared with other state-of-the-art approaches. For Indian Pines and the Pavia University datasets, the results are summarized in Table 1 and Table 2, respectively. It is worth noting that all the classification results were based on the corresponding best parameters.

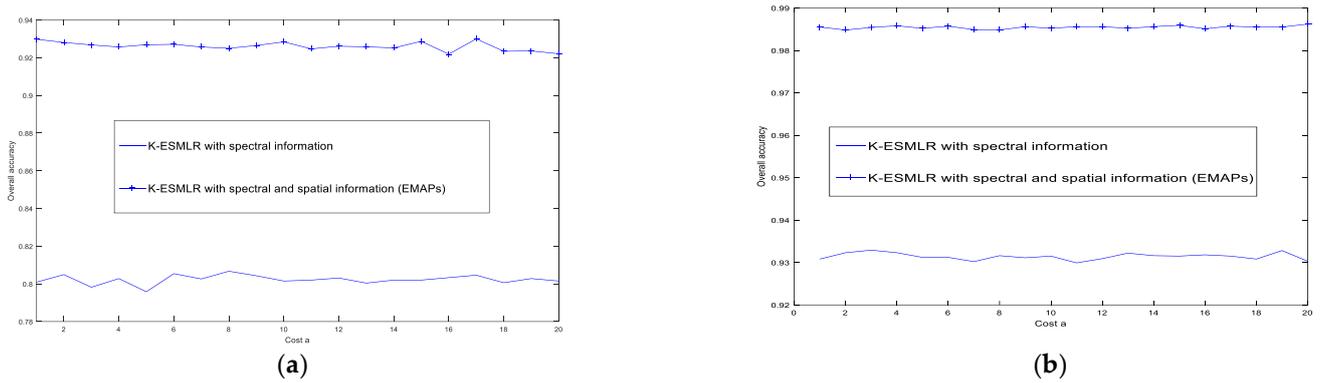

**Figure 4.** The robustness of the proposed framework under different values of cost C: (**a**) The proposed ESMLR with the Indian Pines dataset (using the original HSI dataset, i.e. 200 features, b=-15) and the K-ESMLR with the EMAPs (36 features, b=-17); (**b**) The proposed K-ESMLR with the Pavia University dataset (using the original HSI dataset, i.e. 103 features, b=-10) and the K-ESMLR with the EMAPs (36 features, b=-12).

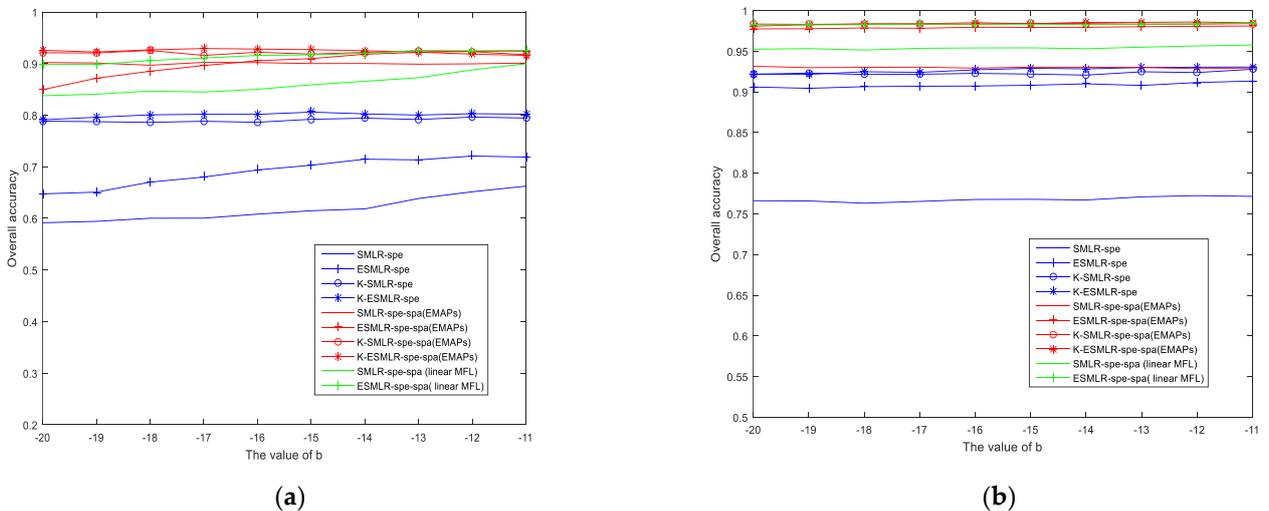

**Figure 5.** The impact of the sparse parameter $b$ ($\lambda = 2^b$) in the proposed framework: (**a**) results on the Indian Pines dataset; (**b**) results on the Pavia University dataset.

From Table 1 and Table 2, it can be concluded that:

1. Compared with the SMLR, the proposed ESMLR achieves better classification results yet the running time are quite comparable under the aforementioned three different situations. Also, it seems ESMLR has a strong learning ability for a small number of training samples when only spectral information is used. For a class with the classification accuracy less than 60%, the improvement is dramatic. This demonstrates the fast and robustness performance of the proposed framework.

2. Compared with K-SVM, they achieve better classification results compared to the ESMLR when only the spectral information is used. However, it requires much more computational time than the proposed approach. When both the spectral and spatial information was used, the proposed ESMLR framework achieves better classification results than K-SVM. This again clearly demonstrates the robustness and efficiency of the proposed framework.

**Table 1.** Classification accuracy (%) with 5% labeled samples in Indian Pines dataset (Best result of each row is marked in bold type).

| | | | Spectral information | | | | | Spectral and spatial information | | | | | | |
| | | | | | | | | EMAPs | | | | | Proposed linear MFL | |
| No | Train | Test | K-SVM | SMLR | K-SMLR | ESMLR | K-ESMLR | K-SVM | SMLR | K-SMLR | ESMLR | K-ESMLR | SMLR | ESMLR |
|---|---|---|---|---|---|---|---|---|---|---|---|---|---|---|
| 1 | 3 | 51 | 60.59 | 2.94 | 25.88 | 19.02 | 41.57 | 88.24 | 86.08 | 89.02 | **89.61** | 88.82 | 86.47 | 88.43 |
| 2 | 71 | 1363 | 78.39 | 75.87 | 76.33 | 75.50 | 77.25 | 86.52 | 85.00 | 86.05 | 87.79 | **90.18** | 86.96 | 89.26 |
| 3 | 41 | 793 | 63.80 | 43.95 | 63.25 | 51.05 | 62.33 | **94.10** | 88.95 | 93.93 | 94.19 | 93.04 | 90.97 | 93.39 |
| 4 | 11 | 223 | 63.72 | 15.47 | 41.66 | 30.18 | 47.89 | 73.10 | 72.02 | 74.98 | 80.76 | 81.57 | 74.62 | **84.39** |
| 5 | 24 | 473 | 88.82 | 72.09 | 87.23 | 82.47 | 87.17 | **91.99** | 88.63 | 89.45 | 89.37 | 91.29 | 90.04 | 90.23 |
| 6 | 37 | 710 | 92.90 | 93.90 | 95.76 | 94.01 | 96.11 | 96.68 | 96.94 | 97.76 | 98.32 | 97.99 | 97.94 | **98.58** |
| 7 | 3 | 23 | 83.04 | 10.87 | 39.13 | 19.57 | 46.09 | **92.17** | 89.13 | 91.74 | **92.17** | 87.83 | 87.83 | 91.74 |
| 8 | 24 | 465 | 97.10 | 99.55 | 99.18 | 99.44 | 98.75 | 99.44 | 98.54 | 99.29 | 97.89 | 99.18 | 99.14 | **99.40** |
| 9 | 3 | 17 | 74.12 | 10.00 | 53.53 | 25.88 | 72.94 | 87.06 | 71.76 | 86.47 | 98.24 | 91.76 | 68.82 | **92.94** |
| 10 | 48 | 920 | 67.57 | 53.68 | 66.97 | 57.84 | 68.11 | 85.73 | 82.21 | 88.28 | **89.47** | 87.70 | 82.87 | 86.85 |
| 11 | 123 | 2345 | 83.20 | 79.14 | 81.16 | 75.94 | 82.39 | 92.51 | 93.26 | 93.74 | 94.29 | 93.27 | 94.52 | **94.84** |
| 12 | 30 | 584 | 74.74 | 52.41 | 74.04 | 61.99 | 79.06 | 81.34 | 75.96 | 83.85 | 84.47 | 85.67 | 81.54 | **91.47** |
| 13 | 10 | 202 | 97.43 | 97.33 | 99.55 | 98.51 | 99.01 | 99.21 | 98.96 | **99.55** | 99.50 | 99.46 | 99.21 | 99.50 |
| 14 | 64 | 1230 | 94.49 | 93.98 | 94.95 | 93.87 | 95.03 | 98.77 | **99.64** | 99.33 | 99.23 | 99.26 | 99.58 | 99.20 |
| 15 | 19 | 361 | 51.25 | 52.96 | 56.87 | 56.98 | 61.55 | 89.67 | 86.29 | 89.09 | 85.35 | 90.03 | 89.31 | **92.66** |
| 16 | 4 | 91 | 78.90 | 66.15 | 54.18 | 66.81 | 57.14 | **97.14** | 87.47 | 81.43 | 75.27 | 84.40 | 79.78 | 82.97 |
| | OA | | 80.55 | 72.61 | 79.13 | 74.46 | 80.35 | 91.50 | 90.04 | 92.01 | 92.46 | 92.75 | 91.43 | **93.44** |
| | AA | | 78.13 | 57.52 | 69.35 | 63.07 | 73.27 | 90.85 | 87.55 | 90.25 | 90.99 | 91.34 | 88.10 | **92.24** |
| | k | | 77.74 | 68.37 | 76.08 | 70.67 | 77.49 | 90.32 | 88.63 | 90.89 | 91.41 | 91.73 | 90.21 | **92.52** |
| | Time(Seconds) | | 7.56 | 0.12 | 0.11 | 0.20 | 0.11 | 2.71 | 0.06 | 0.08 | 0.20 | 0.08 | 0.15 | 0.37 |

**Table 2.** Classification accuracy (%) with 9% labeled samples in Pavia University dataset (Best result of each row is marked in bold type).

| | | | Spectral information | | | | | Spectral and spatial information | | | | | | |
| | | | | | | | | EMAPs | | | | | Proposed linear MFL | |
| No. | Train | Test | K-SVM | SMLR | K-SMLR | ESMLR | K-ESMLR | K-SVM | SMLR | K-SMLR | ESMLR | K-ESMLR | SMLR | ESMLR |
|---|---|---|---|---|---|---|---|---|---|---|---|---|---|---|
| 1 | 548 | 6083 | 90.08 | 73.17 | 88.82 | 87.08 | 90.03 | **98.66** | 90.66 | 97.71 | 98.00 | 98.30 | 96.61 | 97.91 |
| 2 | 540 | 18109 | 94.01 | 78.48 | 94.19 | 92.79 | 94.31 | 97.59 | 93.44 | 98.54 | 98.27 | 98.45 | 96.85 | **99.01** |
| 3 | 392 | 1707 | 84.20 | 70.76 | 83.74 | 79.20 | 85.20 | **97.24** | 83.46 | 96.90 | 96.88 | 97.16 | 93.24 | 96.20 |
| 4 | 542 | 2540 | 97.31 | 94.64 | 97.89 | 97.07 | 97.71 | 99.25 | 97.58 | 98.74 | 98.96 | **99.46** | 98.26 | 99.10 |

| 5 | 265 | 1080 | 99.60 | 99.54 | 99.51 | 99.22 | 99.48 | 99.65 | 99.50 | 99.65 | 98.92 | **99.66** | 99.44 | 99.06 |
|---|---|---|---|---|---|---|---|---|---|---|---|---|---|---|
| 6 | 532 | 4497 | 94.64 | 74.37 | 94.41 | 93.02 | 94.44 | 97.89 | 93.42 | 98.04 | 97.51 | 98.10 | 96.88 | **98.69** |
| 7 | 375 | 955 | 93.93 | 78.23 | 93.03 | 90.43 | 93.60 | 98.01 | 93.51 | 98.27 | **98.57** | 98.42 | 96.19 | 98.45 |
| 8 | 514 | 3168 | 90.32 | 74.72 | 86.09 | 85.83 | 87.85 | 97.90 | 94.86 | 98.01 | **98.29** | 98.25 | 96.49 | 98.10 |
| 9 | 231 | 716 | 99.85 | 96.91 | 99.71 | 99.48 | 99.85 | 99.96 | 99.58 | **99.96** | 99.33 | 99.86 | 99.71 | 99.47 |
| | OA | | 93.21 | 78.50 | 92.72 | 91.28 | 93.18 | 98.46 | 93.23 | 98.30 | 98.17 | 98.44 | 96.83 | **98.61** |
| | AA | | 93.77 | 82.31 | 93.04 | 91.57 | 93.61 | 97.30 | 94.00 | 98.42 | 98.30 | 98.63 | 7.08 | **98.44** |
| | K | | 90.82 | 71.70 | 90.15 | 88.23 | 90.76 | 98.66 | 90.85 | 97.68 | 97.50 | 97.87 | 95.69 | **98.09** |
| | Time (Seconds) | | 106.37 | 0.26 | 3.77 | 0.72 | 3.81 | 53.11 | 0.16 | 3.54 | 0.69 | 3.54 | 0.29 | 1.25 |

*3.5 Classification results with different numbers of training samples*

**Table 3.** Classification accuracy (%) with different numbers of labeled samples in Indian Pines dataset (Best result of each row is marked in bold type).

| | | Spectral information | | | | | Spectral and spatial information | | | | | | |
|---|---|---|---|---|---|---|---|---|---|---|---|---|---|
| | | | | | | | EMAPs | | | | | Proposed Linear MFL | |
| Q | Index | K-SVM | SMLR | K-SMLR | ESMLR | K-ESMLR | K-SVM | SMLR | K-SMLR | ESMLR | K-ESMLR | SMLR | ESMLR |
| 5 | OA | 55.13 | 45.85 | 55.16 | 52.55 | 56.56 | 66.62 | 69.17 | 65.22 | 70.00 | 68.61 | 67.63 | **72.02** |
| | AA | 66.14 | 56.91 | 66.05 | 65.70 | 67.63 | 77.48 | 78.67 | 76.97 | 79.45 | 79.20 | 78.36 | **80.14** |
| | k | 50.02 | 39.57 | 50.17 | 4710 | 51.29 | 62.57 | 65.47 | 61.35 | 66.24 | 64.87 | 63.74 | **68.45** |
| 10 | OA | 63.82 | 52.60 | 62.45 | 57.14 | 63.64 | 74.42 | 78.45 | 78.00 | 79.27 | 77.24 | 79.41 | **80.28** |
| | AA | 75.38 | 64.66 | 74.03 | 71.26 | 74.81 | 84.18 | 85.60 | 86.32 | 86.76 | 85.67 | 86.17 | **87.40** |
| | k | 59.49 | 47.15 | 58.11 | 52.19 | 59.18 | 71.23 | 75.66 | 75.33 | 76.65 | 74.31 | 76.78 | **77.77** |
| 15 | OA | 68.80 | 59.00 | 67.42 | 61.20 | 68.38 | 80.16 | 82.61 | 82.38 | 83.89 | 82.68 | 82.63 | **86.36** |
| | AA | 78.30 | 70.82 | 78.31 | 75.00 | 78.73 | 87.96 | 88.61 | 89.15 | 89.39 | 89.15 | 88.44 | **91.04** |
| | k | 64.95 | 54.20 | 63.49 | 56.59 | 64.47 | 77.61 | 80.34 | 80.12 | 81.75 | 80.43 | 80.34 | **84.53** |
| 20 | OA | 70.55 | 61.30 | 70.66 | 63.65 | 71.06 | 81.51 | 84.98 | 8427 | 87.35 | 85.84 | 85.82 | **87.78** |
| | AA | 80.37 | 71.96 | 80.74 | 75.68 | 82.08 | 89.48 | 89.77 | 91.02 | 91.88 | 91.65 | 90.85 | **92.25** |
| | k | 66.88 | 56.73 | 67.11 | 59.17 | 67.45 | 79.12 | 82.95 | 82.24 | 85.63 | 83.97 | 83.93 | **86.13** |
| 25 | OA | 73.23 | 63.63 | 72.69 | 65.87 | 72.86 | 84.49 | 86.71 | 85.88 | 89.14 | 87.47 | 87.57 | **89.80** |
| | AA | 82.75 | 73.84 | 82.42 | 77.64 | 82.76 | 91.21 | 91.33 | 91.56 | 92.64 | 92.60 | 91.63 | **93.84** |
| | k | .6988 | 59.30 | 69.26 | 61.67 | 69.48 | 82.44 | 84.89 | 84.02 | 87.65 | 85.78 | 85.88 | **88.39** |
| 30 | OA | 73.65 | 65.59 | 73.78 | 66.70 | 75.42 | 86.02 | 88.24 | 87.56 | 89.06 | 88.68 | 88.99 | **89.93** |
| | AA | 83.03 | 75.10 | 73.78 | 78.33 | 84.38 | 91.84 | 91.74 | 92.50 | 93.44 | 93.39 | 92.61 | **93.78** |
| | k | 70.33 | 61.46 | 70.46 | 62.61 | 72.26 | 84.15 | 86.60 | 85.89 | 87.56 | 87.14 | 87.46 | **88.54** |
| 35 | OA | 76.33 | 66.54 | 75.60 | 68.22 | 76.39 | 88.16 | 88.079 | 89.18 | 89.84 | 89003 | 89.84 | **91.02** |
| | AA | 84.75 | 74.99 | 84.10 | 78.83 | 85.29 | 92.87 | 92.41 | 93.71 | 93.74 | 93.70 | 93.26 | **94.43** |
| | k | 73.28 | 62.47 | 72.46 | 64.29 | 73.30 | 86.53 | 87.22 | 87.70 | 88.42 | 87.52 | 88.41 | **89.76** |
| 40 | OA | 77.14 | 67.05 | 75.84 | 69.14 | 77.22 | 89.15 | 89.65 | 89.82 | 90.90 | 90.44 | 90.79 | **91.78** |
| | AA | 85.10 | 75.99 | 84.81 | 79.54 | 85.92 | 93.70 | 92.64 | 94.17 | 94.37 | 94.14 | 93.86 | **94.96** |
| | k | 74.17 | 63.06 | 72.70 | 65.25 | 74.21 | 87.64 | 88.17 | 88.41 | 89.62 | 89.10 | 89.48 | **90.61** |

**Table 4.** Classification accuracy with different numbers of labeled samples in Pavia University dataset (Best result of each row is marked in bold type).

| | Spectral information | Spectral and spatial information | |
|---|---|---|---|
| | | EMAPs | Proposed linear MFL |

| Q | Index | K-SVM | SMLR | K-SMLR | ESMLR | K-ESMLR | K-SVM | SMLR | K-SMLR | ESMLR | K-ESMLR | SMLR | ESMLR |
|---|---|---|---|---|---|---|---|---|---|---|---|---|---|
| 5 | OA | 61.03 | 55.64 | 58.89 | 63.65 | 61.76 | 68.92 | 68.30 | 64.77 | 70.25 | 68.50 | 66.53 | **70.27** |
|   | AA | 72.69 | 64.56 | 70.24 | 71.62 | 72.26 | 77.79 | 69.90 | 73.12 | 72.67 | 77.83 | 72.38 | **76.06** |
|   | k  | 52.24 | 45.47 | 50.14 | 54.72 | 52.93 | 61.28 | 59.73 | 56.63 | 62.30 | 60.84 | 58.64 | **62.60** |
| 10 | OA | 71.25 | 61.81 | 70.41 | 67.93 | 70.73 | 77.35 | 76.18 | 71.19 | 78.17 | 76.93 | 78.07 | **80.98** |
|   | AA | 79.42 | 70.66 | 79.05 | 75.58 | 78.71 | 83.58 | 78.56 | 80.05 | 83.54 | 84.64 | 80.80 | **84.98** |
|   | k  | 63.89 | 52.69 | 62.99 | 60.13 | 63.09 | 71.25 | 69.58 | 64.24 | 72.33 | 70.86 | 71.93 | **75.69** |
| 15 | OA | 74.39 | 64.96 | 72.32 | 73.17 | 73.88 | 82.48 | 77.89 | 79.40 | 81.52 | 78.62 | 80.13 | **84.96** |
|   | AA | 81.50 | 72.84 | 81.93 | 78.56 | 80.79 | 87.37 | 81.79 | 85.05 | 85.88 | 86.38 | 83.67 | **88.31** |
|   | k  | 67.60 | 56.16 | 65.60 | 65.98 | 66.97 | 77.58 | 71.80 | 73.78 | 76.43 | 72.98 | 74.60 | **80.58** |
| 20 | OA | 78.46 | 66.31 | 77.70 | 74.53 | 77.35 | 85.72 | 79.80 | 81.19 | 87.05 | 83.67 | 84.46 | **88.25** |
|   | AA | 83.83 | 73.96 | 83.52 | 79.27 | 83.61 | 89.93 | 83.73 | 87.43 | 89.43 | 88.87 | 87.01 | **90.32** |
|   | k  | 72.47 | 57.80 | 71.65 | 67.58 | 71.16 | 81.59 | 74.27 | 76.14 | 83.18 | 79.02 | 79.94 | **84.71** |
| 25 | OA | 79.56 | 68.93 | 79.12 | 76.15 | 79.53 | 86.54 | 82.43 | 85.08 | 87.35 | 85.81 | 85.71 | **89.59** |
|   | AA | 84.80 | 75.37 | 84.90 | 80.48 | 84.89 | 90.30 | 86.07 | 89.42 | 90.73 | 90.66 | 88.91 | **91.66** |
|   | k  | 73.88 | 60.74 | 73.41 | 69.51 | 73.80 | 82.60 | 77.41 | 80.77 | 83.64 | 81.69 | 81.59 | **86.46** |
| 30 | OA | 80.38 | 69.95 | 80.20 | 77.24 | 80.97 | 88.40 | 84.81 | 85.83 | 89.43 | 88.38 | 87.03 | **90.55** |
|   | AA | 85.54 | 76.28 | 85.50 | 81.32 | 85.87 | 91.59 | 87.63 | 90.62 | 92.27 | 92.03 | 89.98 | **92.34** |
|   | k  | 74.84 | 61.96 | 74.68 | 70.89 | 75.60 | 84.91 | 80.39 | 81.80 | 86.26 | 84.90 | 83.23 | **87.67** |
| 35 | OA | 81.68 | 70.70 | 82.28 | 78.00 | 82.42 | 90.69 | 84.29 | 86.51 | 90.94 | 90.00 | 88.05 | **91.51** |
|   | AA | 85.99 | 76.41 | 86.54 | 82.38 | 87.13 | 93.01 | 88.05 | 91.03 | 92.95 | 92.78 | 90.64 | **93.11** |
|   | k  | 76.40 | 62.73 | 77.26 | 71.87 | 7741 | 87.83 | 79.76 | 82.64 | 88.16 | 86.94 | 84.47 | **88.89** |
| 40 | OA | 83.58 | 70.74 | 83.22 | 78.41 | 83.47 | 90.07 | 85.64 | 87.90 | 91.24 | 90.17 | 89.07 | **92.30** |
|   | AA | 87.25 | 76.85 | 87.27 | 82.16 | 87.41 | 92.83 | 88.48 | 91.98 | 93.46 | 93.19 | 92.11 | **93.56** |
|   | k  | 78.76 | 62.83 | 78.39 | 72.31 | 78.66 | 87.03 | 81.46 | 84.38 | 88.55 | 87.17 | 85.83 | **89.90** |

In this section, we evaluate the robustness of the proposed framework with different numbers of training samples. We vary the number of training samples Q randomly selected from each class, where we have Q=5, 10, 15, 20, 25, 30, 35 and 40 in our experiments. If Q becomes more than 50% of the total samples within a class, only 50% of samples within that class is used for training. For the Indian Pines and the Pavia University datasets, relevant results are summarized in Table 3 and Table 4, respectively.

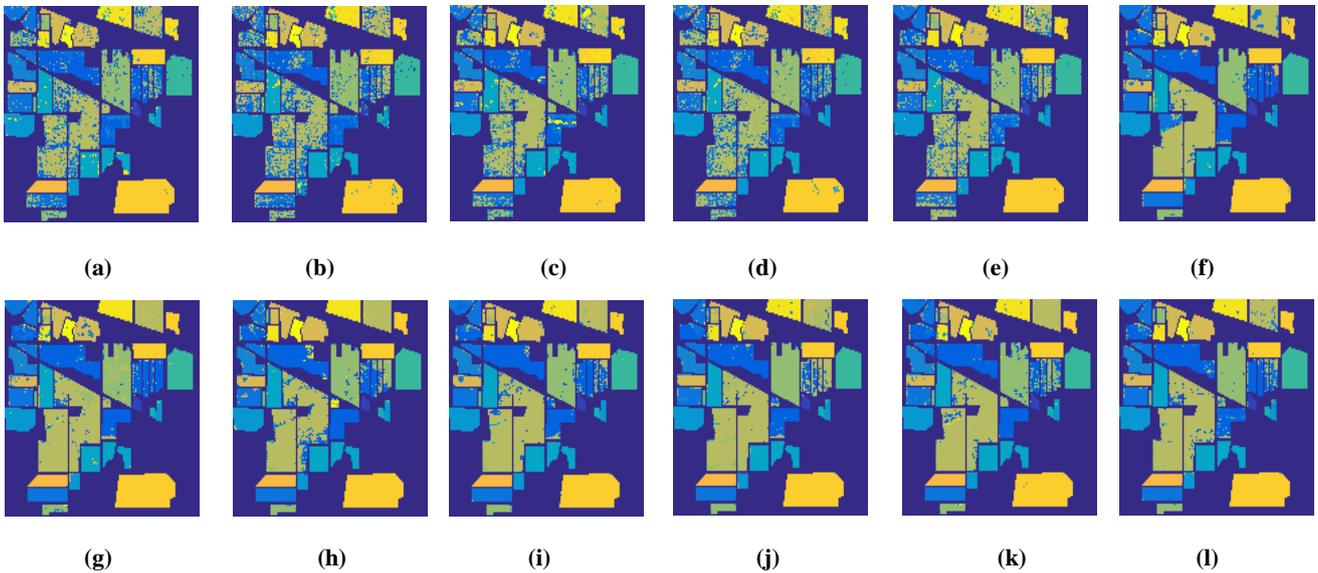

(a) (b) (c) (d) (e) (f)
(g) (h) (i) (j) (k) (l)

**Figure 6.** Results in Indian Pines dataset (with about 40 training samples per class): (**a**) K-SVM (spectral; OA=77.14); (**b**) SMLR (spectral; OA=67.05); (**c**) K-SMLR (spectral; OA=75.84); (**d**) ESMLR (spectral; OA=69.14); (**e**) K-ESMLR (spectral; OA=77.22); (**f**) K-SVM (EMPAs; OA=89.15); (**g**) SMLR (EMAPs; OA=89.65). (**h**) K-SMLR (EMAPs; OA=89.82); (**i**) ESMLR (EMAPs; OA=90.44); (**j**) K-ESMLR (EMAPs; OA=90.44); (**k**) SMLR (proposed linear MFL; OA=90.79); (**l**) ESMLR (proposed linear MFL; OA=91.78).

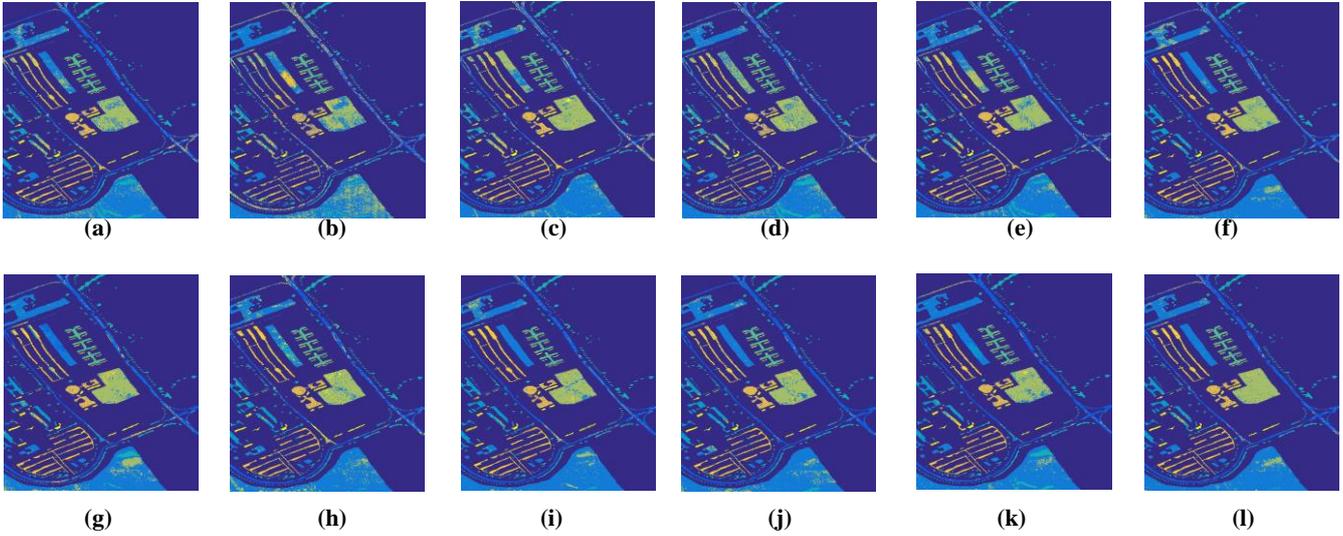

**Figure 7.** Results in Pavia University dataset (with about 40 training samples per class): (**a**) K-SVM (spectral; OA=83.58); (**b**) SMLR (spectral; OA=70.74); (**c**) K-SMLR (spectral; OA=83.22); (**d**) ESMLR (spectral; OA=78.41); (**e**) K-ESMLR (spectral; OA=83.47); (**f**) K-SVM (EMPAs; OA=90.07); (**g**) SMLR (EMAPs; OA=85.64); (**h**) K-SMLR (EMAPs; OA=87.90); (**i**) ESMLR (EMAPs; OA=91.24); (**j**) K-ESMLR (EMAPs; OA=90.17); (**k**) SMLR (proposed linear MFL; OA=89.07); (**l**) ESMLR (proposed linear MFL; OA=92.30);

As seen in Table 3 and Table 4, the proposed ESMLR framework improves the classification accuracy of SMLR dramatically even for a small number of training samples. When both spectral and spatial information are utilized, the proposed framework outperforms K-SVM. As K-SVM requires much more computational time in comparison to the proposed framework, we can conclude that the proposed ESMLR framework provides a fast and robust solution for the classification of HSI.

In addition, Figure 6 and Figure 7 show the classification results from different classifiers for the Indian Pines dataset and the Pavia University dataset, respectively. For each class, the number of training samples is set to 40. Again, this has clearly shown the superior performance of the proposed approach in effective classification of HSI.

## 4. Conclusions

In this paper, we propose a new ESMLR framework to solve the two main drawbacks SMLR for the effective classification of the HSI. By combining linear MFL for incorporating the spectral and spatial information of HSI, the classification accuracy has been successfully improved. Compared with conventional SMLR method, the proposed ESMLR framework has yielded better classification results with a comparable computational time. In comparison to K-SVM, ESMLR requires much less computation time and can better exploit the combination of spatial and spectral information with different labeled numbers of training samples. Furthermore, the proposed approach consistently achieves higher classification accuracy even under a small number of training samples.

The future works will focus on the optimization of the required computational time for K-ESMLR by using sparse representation, and further improvement of the classification accuracy by resorting the ideal regularized composite kernel [47].

**Acknowledgements:** This work is supported by the National Nature Science Foundation of China (nos. 61471132 and 61372173), and the Training program for outstanding young teachers in higher education institutions of Guangdong Province (no. YQ2015057).